\documentclass[11pt]{article}

\usepackage{acl}
\usepackage{times}
\usepackage{latexsym}
\usepackage[T1]{fontenc}
\usepackage[utf8]{inputenc}
\usepackage{microtype}
\usepackage{inconsolata}

\usepackage{booktabs}
\usepackage{multirow}
\usepackage{amsmath}
\usepackage{graphicx}
\usepackage{xcolor}
\usepackage{colortbl}
\usepackage{listings}
\usepackage[normalem]{ulem}
\usepackage{url}

\definecolor{bestcol}{RGB}{223,240,216}
\newcommand{\best}[1]{\cellcolor{bestcol}#1}

\lstdefinestyle{snip}{
  basicstyle=\ttfamily\scriptsize,
  breaklines=true,
  columns=fullflexible,
  keepspaces=true,
  showstringspaces=false,
  frame=single,
  framesep=4pt,
  xleftmargin=2pt,
  aboveskip=4pt,
  belowskip=2pt
}

\title{A Strong Harness Narrows the Model Gap: A Comparative Evaluation for Legislative Document-to-XML Conversion}

\author{Palvika Bansal\textsuperscript{*}, Rittika Jindal\textsuperscript{*}, and Ruichong ``Alex'' Wang\thanks{\ Equal contribution.} \\
  Thomson Reuters Technology \\
  \texttt{\{palvika.bansal, rittika.jindal, alex.wang\}@thomsonreuters.com}}

\begin{document}
\maketitle

\begin{abstract}
Legal-content reproduction requires matching the source exactly in section hierarchy, verbatim
text, citations, and amendment spans; a dropped citation or mislabeled amendment carries into the
products lawyers rely on. We report an industrial pipeline converting U.S. state legislation into
strict, schema-constrained XML, and hypothesize that its accuracy comes from harness engineering,
the deterministic tooling built around a model, more than the model itself. Our controlled
ablation crosses three harnesses (zero-shot, a single tool-using agent, and a decomposed
multi-agent pipeline) with three Claude models on 153 gold bills from two jurisdictions. The
harness narrows the model gap: the weakest model in the full pipeline beats the strongest model
alone (0.948 versus 0.806, paired bootstrap p less than 0.001). Because the two agentic harnesses
share identical tools, their gap comes from orchestration: decomposition plus the persisted state
and per-agent context it brings. Across four scored dimensions, the model spread at zero-shot (0.68 to
0.81) narrows to 0.95 at the full pipeline; the residual sits almost entirely in citation
normalization, where the canonical identifier is retrieved from an external authority rather than
generated from the weights, so a deterministic lookup tool supplies the exactness that a larger model
does not.
\end{abstract}

\section{Introduction}

Most legal-NLP benchmarks reward understanding. Classification, retrieval, judgment
prediction, and summarization all ask whether a system recovers the meaning or the right
label, and a paraphrase that keeps the substance usually counts as correct. Reproduction
is a different problem. Once a bill becomes a product that lawyers rely on, the output has to
match the source down to the surface: sections and subsections in the right order, the bill
text verbatim, citations in canonical form, the struck and inserted spans of an amendment
placed where they belong, and the metadata that indexes it downstream. A dropped citation or a
mislabeled amendment is an error that travels into the products lawyers put in front of clients,
wherever in the document it happens to sit. What we are after is exact reproduction, not a high
average.

A large enough model might be expected to clear this bar on its own. It does not. In the
zero-shot condition (T1), even the strongest model we tested falls short: Opus reaches only
$0.806$ overall and barely resolves jurisdiction-specific citations (normalization $0.232$).
The shortfall is not obviously something a larger model fixes. A model has no way to know a jurisdiction's
normalized statutory identifiers from pre-training, since that mapping lives in an external
authority rather than in the weights, and making the model larger does little to close the gap.

Our working hypothesis is that exactness here owes more to harness engineering, the
orchestration, tools, and deterministic constraints assembled around the model, than to the model
itself. To test it we cross three harnesses with three models (\S\ref{sec:system}): T1 runs the
model zero-shot with no harness, T2 hands a single agent the full tool set and asks it to
self-sequence the conversion, and T3 is the decomposed harness modeled on the production system,
splitting that same tool set across specialized agents. Because T2 and T3 share tools exactly,
whatever separates them should reflect orchestration---decomposition together with the persisted
state and per-agent context it brings---not a wider action space.

We make four contributions. \textbf{(1)~A deployed harness}: a playbook-driven, multi-agent
pipeline that converts U.S.\ state legislation into schema-constrained publication XML and runs
in production at a legal-content publisher (\S\ref{sec:system}). \textbf{(2)~A controlled
model$\times$harness ablation} that holds tool access constant between the single-agent (T2) and
decomposed (T3) harnesses, which lets us attribute the gain to orchestration---decomposition, the persisted
state it introduces, and per-agent context isolation---rather than to a larger action space (\S\ref{sec:results}).
\textbf{(3)~A dimension-resolved error analysis} showing the residual error is not spread evenly
but settles in the legally load-bearing dimensions, citations and amendments, and on the harder
jurisdiction (\S\ref{sec:results}). Handing the citation-lookup tool to a single agent at T2 does
not appear to be enough on its own: only the stronger models invoke it well (normalization $0.88$
and $0.59$ for Sonnet and Opus against Haiku's $0.10$), and the weakest recovers only once T3
gives it a dedicated citation agent ($0.80$, \S\ref{sec:citation}).
\textbf{(4)~An evaluation built for legal-document transformation}: we decompose correctness into
four dimensions, pairing standard document measures (character-level normalized edit distance
(NED) and tree-edit similarity) with two domain measures, exact amendment-span matching and a
citation score split into localization and normalization. That split is what localizes the
residual to citation normalization, which no model in our study resolves unaided (\S\ref{sec:eval}).

\section{Related Work}

\paragraph{Long-document structure and schema generation.}
Systems that convert raw documents into structured markup, such as the OCR-free parsers
Nougat~\citep{blecher2023nougat} and GOT-OCR~\citep{wei2024gotocr}, are judged with the fidelity
metrics we also use: normalized edit distance~\citep{levenshtein1966binary} and Tree-Edit-Distance
Similarity~\citep{zhong2020image}, codified by OmniDocBench~\citep{ouyang2025omnidocbench}. A closer
benchmark, ParseBench~\citep{zhang2026parsebench}, finds no method consistently strong across
dimensions, so general-purpose parsing is not yet solved even before any legal requirement enters.
None of this work, though, scores against a schema with legal-semantic markup such as amendment
spans and resolved citations, at our largest documents' scale (\S\ref{sec:task}), or as a function
of system architecture. Our schema also differs from the legislative-markup standards
USLM\footnote{\url{https://github.com/usgpo/uslm}} and Akoma
Ntoso\footnote{\url{https://www.oasis-open.org/committees/tc_home.php?wg_abbrev=legaldocml}} (the
OASIS LegalDocML standard), since we evaluate against a fixed, confidential in-house schema.
Grammar-constrained decoding~\citep{geng2023grammar} is related but separate: it guarantees
well-formed output at inference time, which is not the same as measuring a full pipeline's fidelity
once assembled.

\paragraph{Agentic harnesses and evaluation.}
Our T3 pipeline draws on now-standard agentic patterns: reasoning interleaved with tool
calls~\citep{yao2022react,schick2023toolformer} and role-specialized
agents~\citep{wu2023autogen,hong2023metagpt}. Benchmarks for such
systems~\citep{liu2023agentbench,jimenez2024swebench,zhou2024webarena} tend to score task
completion and usually vary tool access and decomposition together, so a gain from adding agents is
hard to separate from adding tools. A more recent line names the surrounding infrastructure the
\emph{harness} and argues it, not the base model, often decides whether a capability becomes a
finished task. Harness-Bench~\citep{yao2026harnessbench} reports capability at the
model$\times$harness level, while \citet{zhang2026harness} argue comparisons stay uninterpretable
unless the harness is disclosed and held fixed. We add a domain instance under conditions that line
has not jointly imposed. We grade against a hard exactness bar rather than completion, so the claim
that the harness matters is tested where being almost right already counts as failure. And our
T2-versus-T3 contrast is the fixed-harness comparison \citet{zhang2026harness} ask for, tools held
identical while decomposition varies, which lets the gain read as orchestration rather than a
larger action space.

\paragraph{Legal informatics.}
Applied legal NLP tends to grade a coarser correctness. Benchmarks like
LegalBench~\citep{guha2023legalbench} and CUAD~\citep{hendrycks2021cuad} score classification,
extraction, and QA on label accuracy or span $F_1$, tolerating a near-miss that a legal-publishing
task cannot: a clause classifier that mislabels one clause in twenty is still useful, but a
bill-compilation system that drops one amendment span in twenty misstates an intended change to the
law. Work on statutory citation processing~\citep{darji2026bundesrecht} has treated resolution as a
standalone extraction task. We measure it instead as one dimension of a full pipeline, split into
localization and normalization (\S\ref{sec:eval}). Exact reproduction of a structured legal
document remains comparatively understudied relative to the effort it absorbs.

\section{Task and Dataset}
\label{sec:task}

The task is to convert a legislative bill PDF into structured XML that reproduces the bill
exactly, down to its section hierarchy, verbatim text, statutory citations, and amendment spans, and that validates against a fixed publication schema.

\paragraph{Source material.}
Our inputs are the legislative PDFs that state legislatures publish, here from the Idaho Legislature\footnote{\url{https://legislature.idaho.gov/sessioninfo/}} and the New Jersey Legislature.\footnote{\url{https://www.njleg.state.nj.us/bill-search}} They bring all the difficulty of real legal typography: running headers and footers, margins, multi-column layouts, and footnotes. Amendatory bills add inline formatting on top of this, underline for inserted material and strikethrough for deleted material, which has to be recovered exactly rather than approximately. Text frequently spills across page and column boundaries, so getting the reading order right means reassembling content across physical breaks. Appendix~\ref{app:example} works a real public Idaho bill through the full PDF$\rightarrow$XML pipeline as an example.

\paragraph{Target schema.}
The output has to validate against a strict publication XML schema constrained by a DTD (Document Type Definition), which we keep confidential. It encodes the section hierarchy (title, chapter, section, subsection, with typed paragraph containers), the verbatim bill text, citations resolved to existing statutory identifiers, amendment markup as typed inline elements for added and deleted material, and metadata for indexing. Validity is all-or-nothing, since a single misplaced element can invalidate the whole document. Appendix~\ref{app:example} shows this amendment markup worked through on a real bill.

\paragraph{Dataset.}
The gold set holds 153 expert-produced documents, 103 from Idaho and 50 from New Jersey, spanning the bill revision lifecycle (as-introduced, engrossed, adopted, and amended-substituted versions). Sizes range widely but skew small. The median document has about $130$ body nodes (roughly $5.8$K characters) and a nearly constant tree depth of $8$, since the schema hierarchy is shallow. The tail runs long, out to two New Jersey budget bills large enough to strain the evaluation itself: SB~2025 at 40{,}944
structural nodes and AB~4700 at 8{,}229,\footnote{Both are New Jersey budget
bills; see \url{https://www.njleg.state.nj.us/bill-search/2024/S2025} (SB~2025) and
\url{https://www.njleg.state.nj.us/bill-search/2024/A4700} (AB~4700) for bill text.} both past the node cap of the tree-edit-similarity scorer (\S\ref{sec:eval}). Appendix Table~\ref{tab:corpus} gives the full size distribution by jurisdiction. The two jurisdictions differ in kind. Idaho is a fairly stable baseline, whereas New Jersey is harder across the board, with larger bills, heavier revision lifecycles, and more involved citation patterns. That matters, because New Jersey is where the harness's advantage over the model turns out to be widest (\S\ref{sec:results}).

\section{Harness Engineering: Three Levels of Scaffolding}
\label{sec:system}

\begin{figure*}[t]
  \centering
  \includegraphics[width=0.75\textwidth]{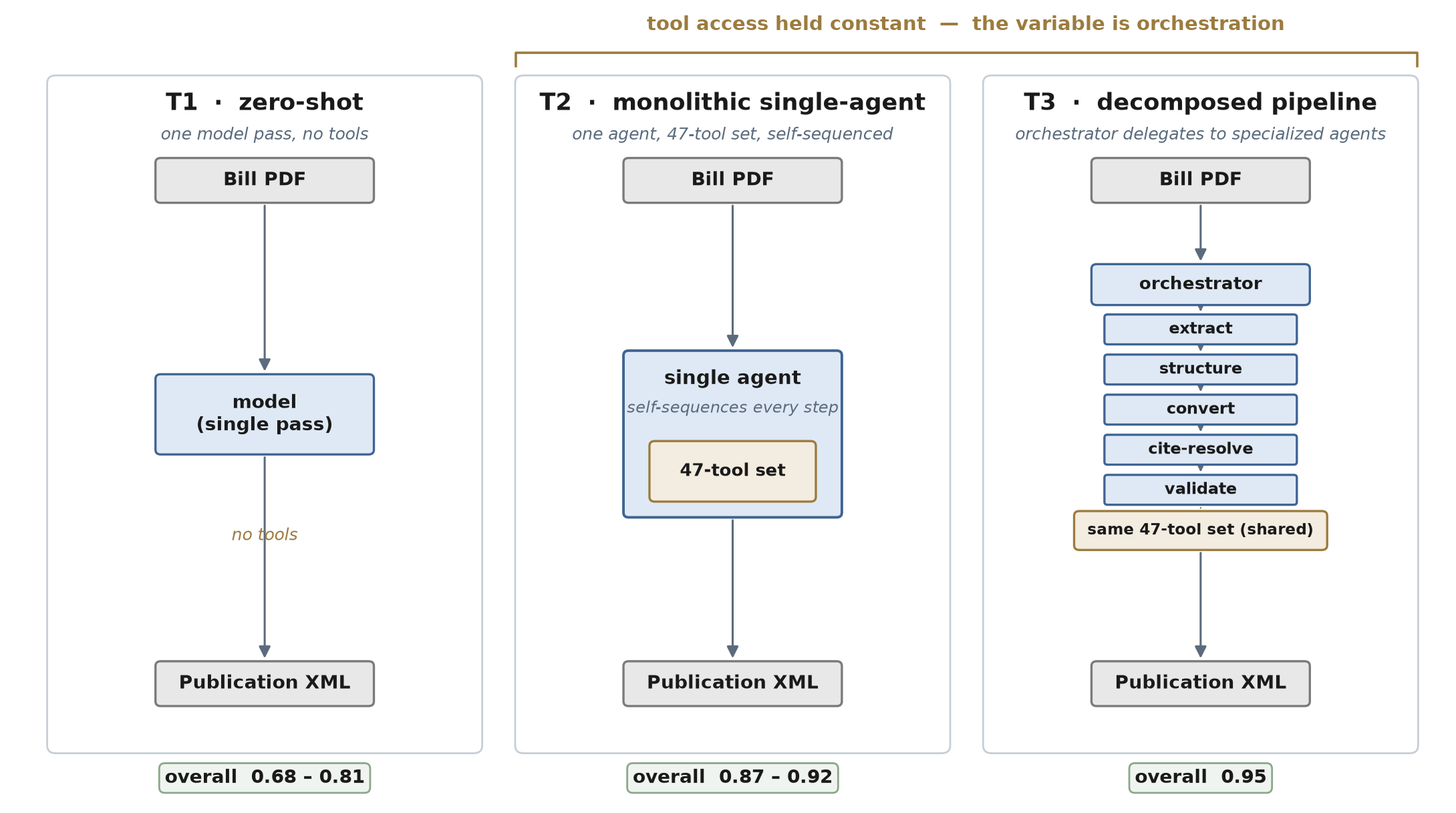}
  \caption{The three harnesses, all starting from the same bill PDF and ending at the same
  publication XML. Because tool access is held constant between T2 and T3, the difference between
  them isolates orchestration: decomposition, the persisted state it enables, and the per-agent
  context it creates.}
  \label{fig:tiers}
\end{figure*}

By harness we mean the deterministic engineering built around the model: typed tools, schema
validation, per-jurisdiction configuration, and structural boundaries that fix what the model
produces at each step. The three harnesses add this scaffolding in increasing amounts, none in
T1, tools in T2, tools plus decomposition in T3 (Figure~\ref{fig:tiers}), and the paper's
question is how much exactness each layer contributes.

\paragraph{T1: no harness, the zero-shot baseline.}
T1 is the model with no harness at all: we give it the source document in one pass and ask for the
publication XML directly, with no tools, retrieval, or intermediate steps. This is the floor the
other harnesses build on. Because it has no legal-code lookup, T1 also cannot resolve a citation to
its canonical identifier, a limitation that dominates the results.

\paragraph{T2: a monolithic harness, the single-agent baseline.}
T2 adds tools but not decomposition, a harness, but an undifferentiated one. It gives one
agent the complete tool set, the same 47
tools the production pipeline uses, and asks that single agent to sequence the whole
conversion itself (extract, structure, convert, resolve citations, validate) in one context
with no sub-agents. T2 is the pivotal control: it shares the model, tools, and data with T3 at
constant tool access, so the T2$\rightarrow$T3 gap isolates \emph{orchestration} rather than a larger
action space. We use that term for three coupled changes T3 introduces together: decomposition into
specialized agents, the persisted checkpointed state that decomposition makes possible, and the
smaller separate context each specialist works in, whereas T2 threads everything through one
long context.

\paragraph{T3: a decomposed harness, modeled on the production pipeline.}
T3 mirrors the deployed system's architecture, with one simplification for experimental control:
every agent runs the same model, whereas the deployed system mixes a stronger orchestrator with
lighter task agents (see the Limitations section). It separates workflow definition from execution. An
editable natural-language playbook, which subject-matter experts change without touching code, is
read by a generic workflow agent that hands each step to a specialized task agent: document
extraction, structure and auxiliary-artifact construction, statute-impact extraction, XML
conversion, citation resolution and footnoting, validation. Each agent therefore works on a narrow,
well-typed subproblem with a small context and action set. The tool set is identical to T2's; only
the orchestration differs. Both harnesses carry state between steps (extracted structure,
intermediate XML, resource and revision identifiers, metadata), and how each handles it is
where the weaker model becomes unstable in T2 (\S\ref{sec:results}). Appendix~\ref{app:arch} gives
the full pipeline and the T2-vs-T3 state-handling contrast.

\begin{table}[t]
  \centering
  \footnotesize
  \setlength{\tabcolsep}{4pt}
  \begin{tabular}{@{}lccc@{}}
    \toprule
    & T1 & T2 & T3 \\
    \midrule
    Bill PDF (native)     & yes  & yes     & yes \\
    XML templates         & yes  & yes     & yes \\
    Jurisdiction rules    & yes  & yes     & yes \\
    \midrule
    Step-ordered playbook & none & 9 steps & 21 steps \\
    Tools                 & none & 47      & 17 \\
    Sub-agents            & none & none    & 7 \\
    Persisted state       & none & none    & yes \\
    \bottomrule
  \end{tabular}
  \caption{What each harness receives. Rows above the middle rule are \emph{knowledge} and are held
  constant; rows below it are \emph{harness}. T2 and T3 reach the same underlying tools: T2 calls
  the 47 leaf tools directly, while T3 reaches those same 47 leaves through 17 composite
  orchestrator tools---the counts differ in exposure, not in access. The step counts differ because T2's playbook stops at the scored artifact, while T3's
  continues into the publication stages that follow it.}
  \label{tab:harness}
\end{table}

\paragraph{What each harness is given.}
Table~\ref{tab:harness} sets out the inputs. All three tiers receive the bill as a native PDF, the
same target-format XML templates, and the same per-jurisdiction conversion rules; T1 carries these
in a single hand-authored instruction, while T2 and T3 load them at run time. The rule we held to
is that only the harness may vary across tiers, never the knowledge, so a capability T2 and T3
reach through a tool is written out in prose for T1 rather than withheld. What T1 cannot be given
this way is anything requiring an external lookup or an intermediate artifact. T2 additionally
receives a nine-step playbook covering the same conversion work T3's playbook describes, so the
step ordering and per-step guidance are not what separates them.

\paragraph{Tool ecosystem.}
Agents reach backend capability over the Model Context Protocol (MCP). One structural property
of it matters for the ablation: MCP is a typed request/response interface with no shared memory
across the boundary. That is exactly what lets us hand the \emph{same} tool set, unchanged, to
both T2 and T3, the invariant the T2-vs-T3 comparison rests on. Six MCP-backed services back the
pipeline: \emph{storage}, \emph{taxonomy}, \emph{ID-mapping}, \emph{NER}, a \emph{codes} service,
and \emph{distribution}, each detailed in Appendix~\ref{app:arch}. One is load-bearing
here: the codes service's \texttt{statute\_lookup\_batch} tool returns each citation's canonical,
jurisdiction-specific identifier in a single round-trip.

\paragraph{The deterministic/model boundary.}
The harness computes deterministically whatever needs no reasoning and leaves the model only
interpretation, and that split is why exactness is a harness property. The codes lookup is the
clearest case: citation resolution is a harness rather than a model capability, because the
canonical identifier is \emph{retrieved} from an authoritative source, not generated from the
weights (\S\ref{sec:citation}). Steps that need no reasoning, such as splitting and converting PDF
pages, validating schema, and generating typed elements from templates, likewise call local
deterministic functions, so what remains for the model is only the part that genuinely requires
interpretation.

\section{Evaluation Framework}
\label{sec:eval}

A single aggregate similarity score cannot distinguish ``reads almost the same'' from ``cites
the wrong law,'' and for this task that distinction is the entire point. The errors that matter tend to be small ones: a section renumbered, a struck word left standing, a citation resolved to the
wrong statute. Any of them can still change what a bill proposes to do. Because a bill is only a \emph{proposed} change, such an error does not corrupt the enacted statute. Its cost lands downstream instead, in how the change gets reported, since acting on a misstated bill can be costly even when the error never reaches the session law. Passing schema validation does not certify any of this: it confirms the document is well-formed and structurally complete (\S\ref{sec:task}), not that its content is right, which is why validity and content correctness have to be scored separately. We therefore break correctness into four independent evaluation dimensions, each scored separately: \textbf{Textual
Integrity}
(the body reproduced without omission, hallucination, or mutation); \textbf{Layout Topology}
(the section/subsection/paragraph hierarchy reconstructed as the correct XML tree);
\textbf{Legislative Mutations} (inline \texttt{added.material}/\texttt{deleted.material} spans
marking exactly which words a bill inserts or strikes); and \textbf{Citation Fidelity}
(statutory cross-references localized in the text and resolved to the canonical identifier).

\paragraph{Textual integrity: normalized edit distance.}
We compute a length-normalized Levenshtein distance over the document body alone (excluding front
matter and identifier fields). Both sides first pass through a normalization function $\phi$ that collapses whitespace, standardizes quotes and dashes. For this metric only, $\phi$ also folds case, since a heading's capitalization is really an artifact of how the source PDF renders it:
\begin{equation}
\mathrm{NED}_{\text{char}}(G,P) = 1 - \frac{\mathrm{Lev}(\phi(G),\phi(P))}{\max(|\phi(G)|,|\phi(P)|)}.
\end{equation}
Read plainly, the score is the share of the document that comes out unaltered: $1.0$ means the two texts are identical once normalized, and $0.9$ means roughly a tenth of the characters would have to be inserted, deleted or changed to turn one into the other.
We report the word-level variant as well, but the composite uses the character-level one, which tends to catch small textual drops that word-level tokenization can hide. Within amendment and citation matching, by contrast, case is \emph{not} folded: a legal mutation or a statute identifier is not a rendering artifact.

\paragraph{Layout topology: tree-edit similarity.}
For structure, we treat the assembled XML as a labeled tree, keeping tags but not text, and score it with Tree-Edit-Distance Similarity (TEDS), a normalized tree-edit distance computed by APTED~\citep{pawlik2016tree} under unit costs:
\begin{equation}
\mathrm{TEDS}(T_G,T_P) = 1 - \frac{\mathrm{TED}(T_G,T_P)}{\max(|T_G|,|T_P|)}.
\end{equation}
Read plainly, this ignores the words entirely and compares only the shape of the document---which
sections hold which subsections and paragraphs---so it asks whether the bill was filed into the
right structural skeleton, however well or badly the text inside it reads.
When either document body is empty or unparseable, we leave TEDS undefined rather than setting it to zero. A missing output should not be quietly recorded as ``completely wrong'' in a measure whose job is to compare two documents that are both actually present.

\paragraph{Legislative mutations: amendment exact-match.}
We score amendment spans by exact multiset match rather than any fuzzy overlap. From gold and prediction we take every \texttt{added.material}/\texttt{deleted.material} span, normalize its text with $\phi$ (this time case-preserving), and treat the two sides as multisets of $(\text{tag},\phi(\text{text}))$ pairs, which gives $\mathrm{TP}=|G\cap P|$, $\mathrm{FP}=|P-G|$, and $\mathrm{FN}=|G-P|$. A document with nothing to mark is not penalized for marking nothing: an empty prediction against an empty gold set counts as a full true negative. We report micro $F_1$ over the documents whose gold actually contains spans, the variant shown in Table~\ref{tab:main}. The composite instead uses an all-documents variant, where a correctly empty document contributes as the true negative just described.

\paragraph{Amendment span placement.}
Matching amendment spans as $(\text{tag},\phi(\text{text}))$ multisets makes the measure above position-blind:
a correctly worded span placed in the wrong location still matches, and in a document
that repeats a sentence the wrong occurrence could be marked. To bound the effect, we
re-scored with each span's preceding in-paragraph context appended to the match key.
Of the spans the reported metric credits, $88.1\%$ also agree on context, so at most
$11.9\%$ could be mislocated---an upper bound, since an unrelated edit elsewhere in
the same paragraph also breaks the context key. Adding the context requirement lowers
mean amendment $F_1$ by $0.082$, $0.085$ and $0.094$ at T1, T2 and T3 respectively---
nearly the same penalty at every level of scaffolding---so position-blindness does not
differentially favor any harness, and the ordering in \S\ref{sec:results} is unchanged.

\paragraph{Citation fidelity: localization vs.\ normalization.}
Citations use the same multiset-F machinery, split into two independent signals over \texttt{cite.query} nodes. \emph{Localization} asks whether the system found and tagged the citation text inside the node at all. \emph{Normalization} asks whether the citations it did find were resolved to the correct canonical identifier, matching on that identifier rather than on the visible text. The two fail separately: a system may localize perfectly yet resolve nothing, and \S\ref{sec:results} shows this split is where the model-versus-harness distinction is sharpest. Collapsing them into one $F_1$ would score a found-but-unresolved citation the same as one never found. We report micro $F_1$ as the headline. One caveat: because a match requires the citation to be wrapped in \texttt{cite.query}, localization partly reflects harness-supplied schema-tagging competence, not citation-finding on its own.
Two consequences follow from matching multisets. Repeated citations are counted rather than
collapsed: if a statute is cited $k$ times in the gold document, the prediction must carry $k$
matching instances, and any surplus or shortfall scores as a false positive or false negative
like any other. Unresolved citations are handled differently by the two signals. A span the
system tagged but left unresolved still counts as found for localization, since it was in fact
located; it contributes no identifier to normalization, so there it is a miss against the gold
identifier rather than being excused. A citation the gold itself leaves unresolved is dropped
from both sides and is not required of the system.

\paragraph{Composite.}
We fold text and structure into a single headline number, the unweighted mean of character-level NED, TEDS, and amendment $F_1$:
\begin{equation}
\mathrm{Overall} = \tfrac{1}{3}\left(\mathrm{NED}_{\text{char}} + \mathrm{TEDS} + \mathrm{Amend.}\ F_1\right).
\end{equation}
Citation localization and normalization stay out of this composite and are reported on their own, since averaging them in would let near-perfect text and structure mask a citation failure. TEDS is worst-case cubic in the number of nodes, so we cap it at 4{,}000 nodes and leave TEDS, and with it Overall, unscored for the two largest and hardest documents (Appendix Table~\ref{tab:stress}). Those two drop out of the composite, though every other dimension is still computed for them. Unparseable files and output-integrity failures, whether a truncated or an over-generated output, are recorded as a distinct class rather than allowed to abort the cell. In the event every output we scored parsed as well-formed XML and none was rejected as malformed.

\paragraph{Statistical testing.} Every cell mean in \S\ref{sec:results} carries a 95\% confidence interval from a bootstrap over that cell's scored documents (10{,}000 resamples, fixed seed). For the comparisons our claims rest on, such as Haiku$\cdot$T3 against Opus$\cdot$T1 or Sonnet against Opus within a tier, we run a paired bootstrap over documents scored in both arms, matched by identity, and report a two-sided $p$-value on the mean difference. We prefer this to a chi-squared test, since our scores are continuous on $[0,1]$ rather than pass/fail counts and binning them would discard information the bootstrap keeps. This captures variance across the \emph{document sample}, not run-to-run variance, as each cell is a single run. We leave the $p$-values uncorrected, but the confirmatory comparisons form a small pre-specified family of about six, and every effect we rely on reaches $p\le0.004$, below a Bonferroni threshold for that size ($\alpha/6\approx0.008$). The one contrast that misses significance, Sonnet against Opus at T2 ($p=0.39$), we decline to interpret.

\section{Experimental Results}
\label{sec:results}

Each harness was run on the same 153-document cohort across the three models: 459 attempted runs
per harness, the full two-jurisdiction cohort of \S\ref{sec:task}.\footnote{Model API identifiers:
\texttt{claude-haiku-4-5-20251001}, \texttt{claude-sonnet-4-6}, \texttt{claude-opus-4-6}; these
name the exact model snapshots we ran (see the Limitations section). No temperature override or
seed is set (provider default), and each (model, harness, document) cell is a single run, not
averaged over repeats. Haiku, Sonnet and Opus are the vendor's capability tiers; parameter counts are not published, and the tiers also differ in context window and maximum output.}
Cells are scored over the runs that yielded a scorable document.

\begin{table*}[t]
  \centering
  \small
  \setlength{\tabcolsep}{7pt}
  \begin{tabular}{llccccccc}
    \toprule
    & & & \multicolumn{3}{c}{Structure \& text} & & \multicolumn{2}{c}{Citations} \\
    \cmidrule(lr){4-6}\cmidrule(lr){8-9}
    Model & Harness & Overall $\uparrow$ & NED $\uparrow$ & TEDS $\uparrow$ & Amend.\ $F_1$ $\uparrow$ & & Local.\ $\uparrow$ & Norm.\ $\uparrow$ \\
    \midrule
    \multirow{3}{*}{Haiku 4.5}
      & T1 & 0.677 {\scriptsize[.654,.700]} & 0.846 & 0.647 & 0.288 & & 0.206 & 0.024 \\
      & T2 & 0.872 {\scriptsize[.816,.921]} & 0.882 & 0.869 & 0.808 & & 0.367 & 0.103 \\
      & T3 & \best{0.948} {\scriptsize[.932,.963]} & \best{0.982} & \best{0.946} & \best{0.876} & & \best{0.813} & \best{0.801} \\
    \midrule
    \multirow{3}{*}{Sonnet 4.6}
      & T1 & 0.807 {\scriptsize[.785,.829]} & 0.872 & 0.794 & 0.598 & & 0.185 & 0.282 \\
      & T2 & 0.922 {\scriptsize[.899,.943]} & 0.948 & 0.915 & 0.871 & & 0.869 & 0.881 \\
      & T3 & \best{0.954} {\scriptsize[.937,.969]} & \best{0.983} & \best{0.954} & \best{0.892} & & \best{0.954} & \best{0.954} \\
    \midrule
    \multirow{3}{*}{Opus 4.6}
      & T1 & 0.806 {\scriptsize[.778,.834]} & 0.858 & 0.789 & 0.653 & & 0.229 & 0.232 \\
      & T2 & 0.910 {\scriptsize[.879,.937]} & 0.936 & 0.904 & 0.847 & & 0.631 & 0.593 \\
      & T3 & \best{0.946} {\scriptsize[.916,.970]} & \best{0.985} & \best{0.955} & \best{0.898} & & \best{0.943} & \best{0.943} \\
    \bottomrule
  \end{tabular}
  \caption[Main results: per-dimension and composite exactness across model$\times$harness cells.]{Main results: per-dimension and composite exactness for all nine model$\times$harness cells, pooled across both jurisdictions (micro $F_1$ for citations and amendments, higher is better throughout). Every model climbs monotonically from the zero-shot baseline (T1) to the full pipeline (T3), and the full-pipeline rows are shaded. \texttt{Overall} is the unweighted mean of NED, TEDS, and amendment $F_1$. The two citation columns are reported separately and excluded from it (\S\ref{sec:eval}). Because it averages per document over a slightly different population than its displayed columns, \texttt{Overall} need not equal their row-wise mean (see footnote). Bracketed ranges beneath \texttt{Overall} are 95\% bootstrap confidence intervals (\S\ref{sec:eval}).\footnotemark}
  \label{tab:main}
\end{table*}
\footnotetext{\texttt{Overall} need not equal the mean of this table's own NED/TEDS/Amend.\ $F_1$
  columns: it is computed per document then macro-averaged, and each column pools over a different
  document subset (TEDS drops the two capped documents; the displayed Amend.\ $F_1$ is the
  applicable-only variant, whereas \texttt{Overall} uses the all-documents variant). The gap is
  largest at zero-shot; Appendix~\ref{app:detail} works the arithmetic through.}

\begin{figure}[t]
  \centering
  \includegraphics[width=\columnwidth]{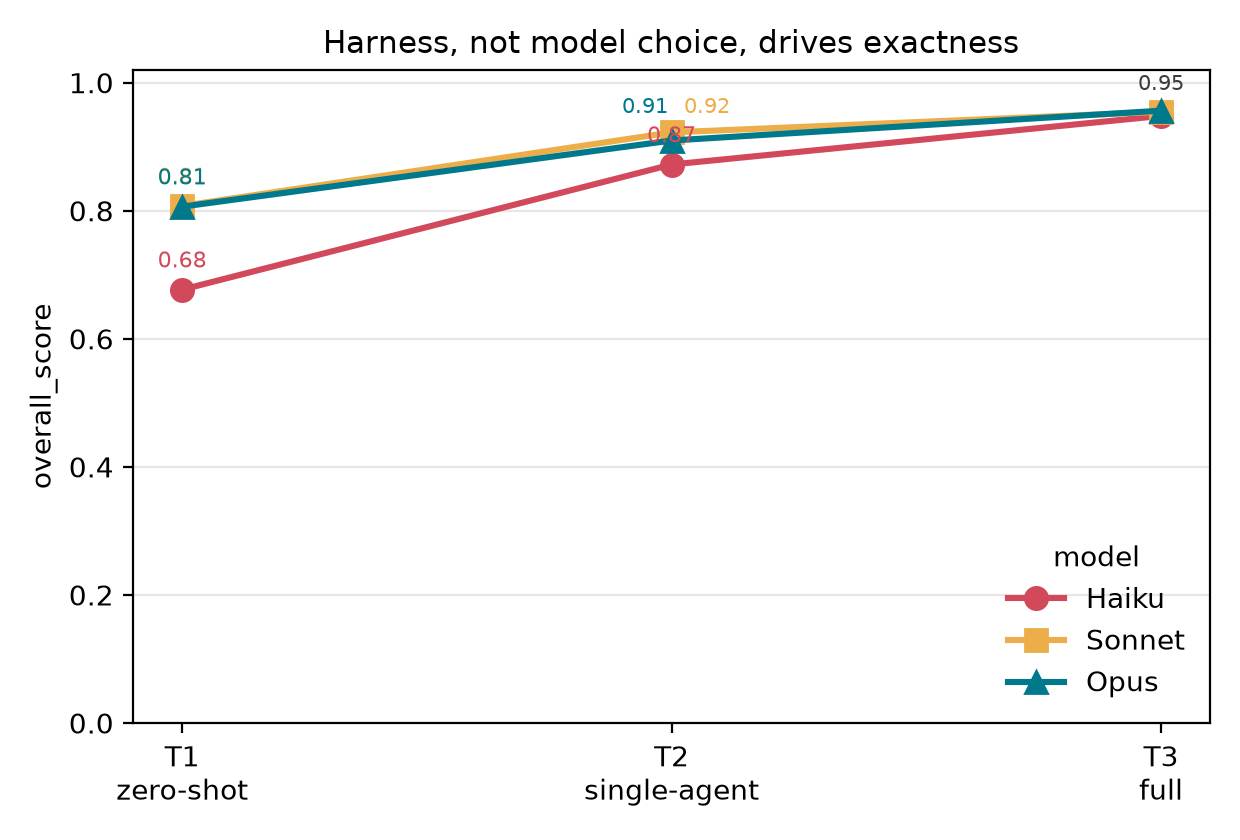}
  \caption{\texttt{Overall} exactness rises with the harness for every model. Unaided, the models spread widely, but inside the full system they converge.}
  \label{fig:staircase}
\end{figure}

\subsection{The staircase: harness dominates model}
\label{sec:staircase}

Table~\ref{tab:main} reports every dimension and the headline \texttt{Overall} score for all nine cells, plotted in Figure~\ref{fig:staircase}. The staircase is clean and monotonic for every model, and the model gap shrinks at every step. Moving Haiku from zero-shot to the full system raises \texttt{Overall} from $0.677$ to $0.948$, whereas moving from Haiku to Opus at zero-shot raises it only from $0.677$ to $0.806$: the harness lifts more than the model does. Unaided, the models span $0.677$--$0.807$, and in the full system $0.946$--$0.954$. The comparison the ablation isolates holds under a paired test: Sonnet$\cdot$T3 ($0.954$) and Opus$\cdot$T3 ($0.946$) stay close, and both clearly exceed Opus$\cdot$T1 ($0.806$, paired bootstrap $\Delta=+0.120$, $p<0.001$, $n=124$). Even Haiku$\cdot$T3 ($0.948$) beats the strongest model zero-shot ($\Delta=+0.117$, $p<0.001$, $n=133$).

One result complicates the ``Haiku~$<$~Sonnet~$<$~Opus'' ranking. At T2, Sonnet's \texttt{Overall} ($0.922$) edges out Opus's ($0.910$), a gap a paired bootstrap cannot distinguish from noise ($p=0.39$, $n=151$). What is robust is the dimension underneath. The gap is almost entirely in citations, where Sonnet resolves far better than Opus at T2 (localization $0.869$ vs.\ $0.631$, normalization $0.881$ vs.\ $0.593$, at $p=0.004$ and $p<0.001$, $n=74$), while structure and text are near-tied, so the composite dilutes a 24-to-29-point (on the 0--1 scale) citation gap below significance. The reversal is specific to T2, so tier alone does not predict which model best exploits a harness. The model axis reads as weak (Haiku) versus strong (Sonnet and Opus, trading the lead) rather than a clean three-point ranking.

\subsection{Per-dimension: what the harness fixes}

The composite hides structure. Across the five dimensions (Table~\ref{tab:main}, Appendix Figure~\ref{fig:dimensions}) the split is clear: text and structure are close to solved once any tool-using harness is present, with NED and TEDS at $.95$--$.99$ by T3 and much banked at T2, while amendments improve mostly on the T1$\rightarrow$T2 jump before flattening. Citations are the sharpest story, treated next.

\subsection{Citation resolution is a harness property}
\label{sec:citation}

\begin{figure}[t]
  \centering
  \includegraphics[width=\columnwidth]{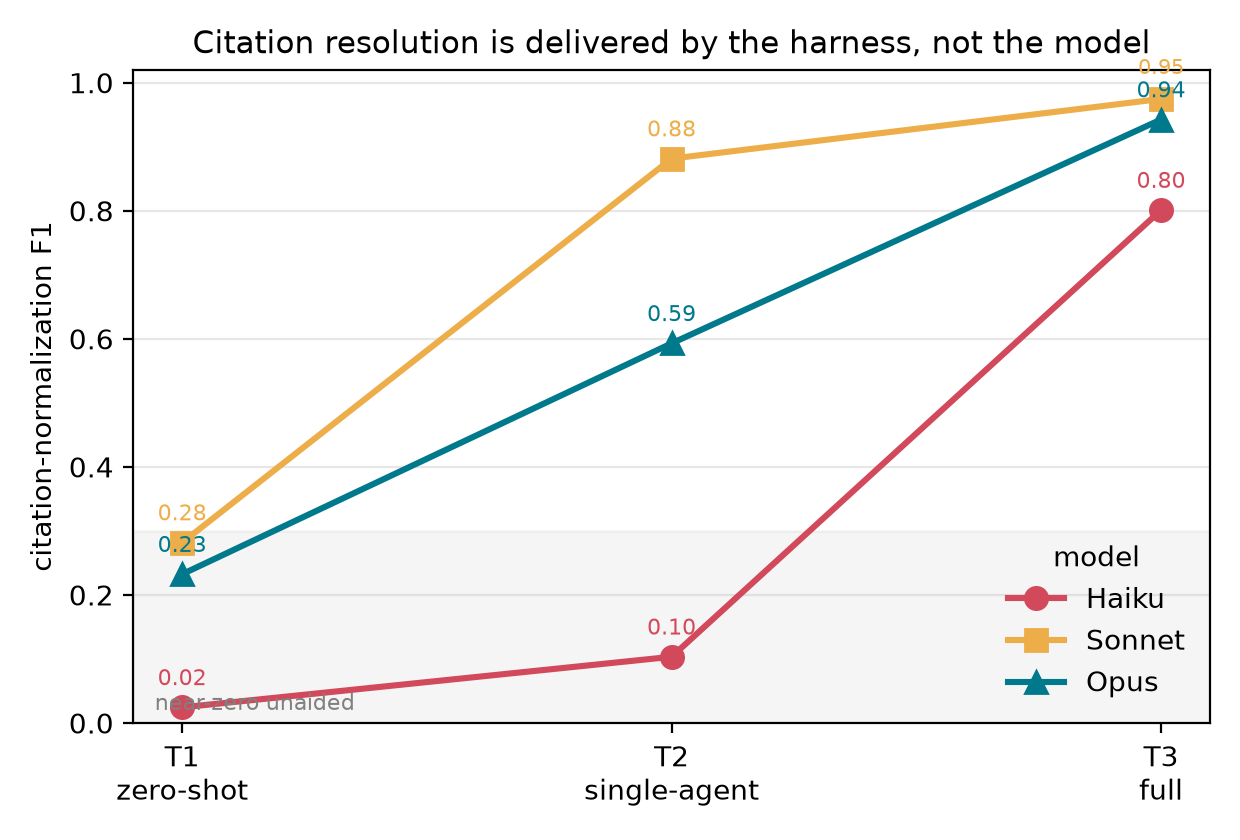}
  \caption{Citation normalization is near-zero for every model unaided and is delivered by the harness. The stronger models resolve citations as soon as they have the tool (T2). The weakest model does not, and reaches parity only under full decomposition (T3).}
  \label{fig:citation}
\end{figure}

The sharpest split in the model$\times$harness interaction shows up here (Figure~\ref{fig:citation}). Haiku goes from near-zero (T1 $.024$) to resolving citations only once the harness is fully decomposed (T2 $.103$, T3 $.801$), whereas Sonnet and Opus resolve well as soon as they have the lookup tool at all (T2 cite-norm $.881$ and $.593$, up from T1 $.282$ and $.232$). The same tool gives the strong models a large jump and the weak one almost none until decomposition is added. We read that as the dedicated citation agent substituting for a capability Haiku lacks rather than merely adding one, though the measurement shows only the gap, not its cause. Localization is separable from normalization, not its easier half: at T1 Sonnet localizes \emph{worse} than Haiku ($.185$ vs.\ $.206$) yet normalizes far better once a citation is found ($.282$ vs.\ $.024$). Since T1 has no lookup tool by construction, its near-zero normalization floor is not the finding. What matters is how unevenly that floor is escaped once the tool arrives, and only for the model that also gets a dedicated agent.

To characterize what is left, we diffed predicted against gold \texttt{cite.query} resolutions for every T3 output with imperfect normalization (32 of 167 outputs containing at least one \texttt{cite.query} node, pooled across the three T3 models, detailed in Appendix Table~\ref{tab:citetax}). The dominant mode is a \emph{dropped subset}. Most citations resolve, but a group is silently omitted, typically those in a statutory title the resolution step never reaches (NJ~SB~3587). Rarer are granularity errors (ID~SB~1023), lookup misses (ID~SB~1142), and one over-expansion outlier that alone inflates Idaho's raw residual (ID~SB~1008). The dominant modes look coverage-limited rather than reasoning-limited: they are what we would expect when the codes service returns no identifier, the same per-jurisdiction dependency we flag as a generalization risk (see the Limitations section).

\subsection{Reliability: the tail collapses}

Means understate the case for the harness. What matters is how often it produces a badly wrong document, and any inaccuracy is costly whether or not it is substantive, so the risk to manage is the low-scoring tail, not the average. Table~\ref{tab:reliability} reports the pooled variance and the rate of low-scoring documents ($<0.70$) by harness. Variance is non-monotonic: T2 is the widest, not the narrowest, before the full pipeline collapses it. At T1 the low-score rate splits sharply by model ($52\%$ Haiku, $23\%$ Opus, $19\%$ Sonnet), then falls, pooled across models, to $8\%$ at T2 and $4\%$ at T3. Haiku$\cdot$T2 is the single most unstable cell (std $0.248$). That is the measurement; our explanation for it is that a weak model asked to self-sequence is erratic, and that what steadies it is orchestration rather than a stronger model. The per-document distribution behind these statistics appears in Appendix Figure~\ref{fig:reliability}.

\begin{table}[t]
  \centering
  \small
  \begin{tabular}{lcc}
    \toprule
    Harness & Pooled std $\downarrow$ & Low-score ($<0.70$) $\downarrow$ \\
    \midrule
    T1 & 0.167 & 31\% \\
    T2 & 0.188 & 8\% \\
    T3 & \best{0.099} & \best{4\%} \\
    \bottomrule
  \end{tabular}
  \caption{Pooled variance and rate of low-scoring documents ($<0.70$), by harness. The full
  pipeline (T3) both tightens the distribution and shrinks the tail.}
  \label{tab:reliability}
\end{table}

Before the final numbers, we sanity-checked every output for integrity and re-ran flagged cases to tell transient errors from stable failures. This caught 7 of 459 T1 attempts ($1.5\%$) and 21 of 459 T2 attempts ($4.6\%$), concentrated on the largest New Jersey bills, and repeated runs reproduced them, so we record them as stable properties of the harness rather than transient noise. That T2 fails more often than T1 is the observation; the explanation we offer is that it asks one model to self-sequence extraction, conversion, lookups, and validation in a single context, spending output and context budget on orchestration. The failures run both ways. AB~3678 (302k gold chars) truncates to $\approx$4.7k at T2, while AB~4878 over-generates to $\approx$2.1$\times$ gold for all three models identically. T3, sharing the same tool access, avoids both.

\subsection{Cross-jurisdictional stress testing}

\begin{table}[t]
  \centering
  \small
  \setlength{\tabcolsep}{4pt}
  \begin{tabular}{lcccccc}
    \toprule
    & \multicolumn{3}{c}{Idaho} & \multicolumn{3}{c}{New Jersey} \\
    \cmidrule(lr){2-4}\cmidrule(lr){5-7}
    Harness & Haiku & Sonnet & Opus & Haiku & Sonnet & Opus \\
    \midrule
    T1 & 0.695 & 0.833 & 0.860 & 0.636 & 0.754 & 0.697 \\
    T2 & 0.917 & 0.972 & 0.964 & 0.783 & 0.816 & 0.798 \\
    T3 & \best{0.974} & \best{0.977} & \best{0.976} & \best{0.879} & \best{0.871} & \best{0.809} \\
    \bottomrule
  \end{tabular}
  \caption{\texttt{Overall} score by jurisdiction and harness; New Jersey is harder under every
  harness, yet the harness's advantage over the unaided model is \emph{wider} there, not narrower.
  Full-pipeline (T3) cells shaded. Each cell averages over its own scored documents, so the
  jurisdiction columns combine into Table~\ref{tab:main}'s \texttt{Overall} weighted by those
  counts, not by the cohort totals.}
  \label{tab:juris}
\end{table}

New Jersey is harder under every harness and for every model (Table~\ref{tab:juris}). More importantly, the thesis holds \emph{more} strongly on the harder jurisdiction: the gap between Sonnet in the full system and Opus unaided (Sonnet$\cdot$T3 $-$ Opus$\cdot$T1) is $+0.117$ on Idaho and $+0.174$ on New Jersey. Were the harness's advantage merely an artifact of Idaho being easy, that gap should shrink on the harder cohort, but it widens. The two New Jersey budget bills make the point (gold trees of 8{,}229 and 40{,}944 nodes): at zero-shot the model abandons both, while the full pipeline reconstructs them to near-solved scores, both past the TEDS node cap (per-bill numbers in Appendix Table~\ref{tab:stress}).

\section{Conclusion}

We asked whether exactness here is bought with a bigger model or a better harness, and our
experiments lean firmly toward the harness. Scores climb a clean staircase, and the weakest model
in the full pipeline beats the strongest model unaided (Haiku$\cdot$T3 $0.948$ against
Opus$\cdot$T1 $0.806$). Because T2 and T3 keep tools constant, the further climb comes from
orchestration---decomposition, the state persistence it brings, and per-agent context---not from
more tooling. The pipeline also draws
very different models together, tightens the distribution, thins the low-scoring tail, and supplies
the dimension that carries the most legal weight, citation normalization, which no model reaches
unaided. The reason is specific: a citation's canonical identifier comes from a lookup against an
external authority, not from the weights, so a deterministic tool plus the decomposition that
reliably invokes it delivers what a larger model alone does not (\S\ref{sec:citation}).
Every result here comes from a single model family (Claude Haiku~4.5, Sonnet~4.6
and Opus~4.6); whether the harness effect transfers to other model families is
untested, and we leave it to future work.

What remains is small and specific: citation exactness and the harder New Jersey documents, where
the last bit of accuracy still tracks the model. A larger model is an unlikely fix: the identifier
has to be retrieved from an external authority, not generated. The residual is also the part of the
task carrying the most legal risk, where being almost right is not good enough. This is where human judgment belongs: expert review is a permanent part
of the architecture, not a temporary crutch. Because the residual is concentrated and identifiable,
it can be routed, with the system auto-publishing what it has solved and surfacing the citation
and amendment decisions that carry legal weight for an editor to confirm. The chief open problem is confidence estimates calibrated well enough to drive that routing. The
same pattern likely holds wherever an exactness bar meets an external source of truth that
pretraining cannot supply, as in medical coding or regulatory tagging: wire that lookup in as a
deterministic tool, decompose so it is reliably invoked, and route the residual to expert review.

\section*{Limitations}

\paragraph{What we score.}
We score the pre-review XML, captured after the pipeline's automated validation step but before
the editorial-review gate (Appendix~\ref{app:arch}): the system's final unaided output, not the
human-corrected artifact that reaches publication. The T3 cells
also use a uniform model across every agent, whereas the deployed system is mixed (a stronger
orchestrator with lighter task agents); this is the controlled condition, not the production one.

\paragraph{Generalization.}
We study two jurisdictions. Nothing in the harness is jurisdiction-specific by construction:
formatting, citation conventions, and schema mappings enter as per-jurisdiction configuration
(\S\ref{sec:system}) rather than code, so the architecture is \emph{intended} to generalize to the
remaining U.S. states. Having tested only two, we read these results as a two-jurisdiction,
single-model-family case study. Four concrete risks
could break the claim elsewhere: amendment conventions other than strikethrough/underline;
citation and section-numbering formats each needing their own normalization mapping; source-PDF
digitization worse than Idaho's or New Jersey's; and codes-lookup coverage that may not yet exist
for a new state. All results use
Claude-family models (Haiku 4.5, Sonnet 4.6, Opus 4.6); whether the harness effect holds for other
model families is untested and left to future work.

\paragraph{One document type.}
We convert legislative bills under a single publication schema. Treaties, judgments, and
regulatory codes demand the same exactness but carry their own conventions and their own notions
of what an amendment or a citation is, and we have not tested them. We expect the
finding to transfer, since it turns on wiring an external authority in as a deterministic tool
rather than anything bill-specific, but that is an expectation, not a result.

\paragraph{What the T2--T3 contrast isolates.}
The comparison holds tool access fixed, but not everything else. Alongside decomposition, T3 adds
persisted, checkpointed state that T2 lacks (Appendix~\ref{app:arch}) and gives each specialized
agent its own small context, whereas T2 re-threads every intermediate result through one long
context. These three travel together by construction, so we read the T2$\rightarrow$T3 gain as
orchestration rather than decomposition alone; ablations separating them are left to future work.

\paragraph{What we do not measure.}
We report no tokens, latency or dollar cost per setting: the ablation runs were not instrumented
for per-run usage accounting, and the figures cannot be recovered from the artifacts we kept. The
one indication we can give is a configuration bound, not a measurement: the single-agent tier
needed a far higher per-run request ceiling than the orchestrator, since its request count grows
with bill length. We report no schema-validity rate either, because DTD validation
runs in the downstream publish step rather than on the pre-review artifact we score and so sits
outside our scorer suite. Both a cost-per-document comparison and a uniform validity rate across
the nine cells are left to future work.

\paragraph{Metrics and gold set.}
The reported cells score output-integrity failures---truncated or over-generated outputs
(\S\ref{sec:results})---as produced rather than setting them aside, on the view that a harness
which truncates a document has failed the task. The consequence is that the T1 and T2 deficits
blend conversion quality with output robustness, and we do not report the two components
separately; the counts are small and concentrated (7 of 459 at T1, 21 of 459 at T2). Our metrics are exactness proxies: high scores
are necessary for legal correctness but not sufficient. The gold set comes from a legacy
production system---unrelated in architecture, prompts, and tooling to the pipeline evaluated
here---then corrected by a subject-matter expert against the source bill; sharing no lineage with
any of the three tiers, it advantages none differentially. Because that correction is a single-SME
review, we report neither inter-annotator agreement nor a human-performance ceiling; both would
require a second expert.

\paragraph{Reproducibility.}
The gold set is proprietary and the DTD is confidential, so we name the mitigations we do have.
The source inputs are public government documents (the Idaho and New Jersey legislatures,
\S\ref{sec:task}) and every bill we single out is cited by its public number, so each example's
PDF side is independently retrievable; Appendix~\ref{app:example} works one public bill end to end.
All scorers are standard and deterministic (Levenshtein NED, APTED tree-edit similarity, multiset
$F_1$) run with a fixed seed (\S\ref{sec:eval}), and the three models are pinned by exact API
snapshot (\S\ref{sec:results}; for the 4.6 generation the dateless identifier \emph{is} the pinned
snapshot, not an evergreen alias). Running the harness on a public document-parsing benchmark
would give our text and structure dimensions an external anchor, and we leave that to future
work.

\paragraph{Baselines.}
All our baselines are LLM-based: T1 is a zero-shot LLM floor, not a tuned non-LLM parser such as an
OCR-plus-rules pipeline or the document-AI systems we cite, so our results show the harness lifts a
weak LLM to exactness, not that it outperforms a competent deterministic parser; benchmarking such
a parser on this data is left to future work.

\section*{Ethical Considerations}

The system operates where errors can carry legal consequences, so our results are as much a caution as an endorsement of automation. The residual error concentrates in the dimensions that carry legal meaning, citations and amendments, which is exactly where over-trust would do the most harm. Human review of those decisions is therefore not an optional refinement but a requirement for responsible deployment. The evaluation uses expert-produced gold and involves no personal data.

\bibliography{references}

\appendix

\section{Worked PDF$\rightarrow$XML Example}
\label{app:example}

Bills are public government works; the excerpt below is from Idaho House Bill No.\
66 (2025),\footnote{\url{https://legislature.idaho.gov/sessioninfo/2025/legislation/H0066/}}
which amends Idaho Code \S\,72-713. It shows the amendatory structure the pipeline has to recover, with strikethrough for deleted material and underline for inserted material, alongside a representative fragment of the publication XML (element names are stand-ins, since the schema vocabulary is confidential). In that fragment the inserted and struck spans become typed inline elements, and the statutory reference becomes a resolved \texttt{cite.query}.

\begin{center}
\fbox{\parbox{0.92\columnwidth}{\ttfamily\scriptsize
SECTION 1. That Section 72-713, Idaho Code, be, and the same is hereby amended to read as follows:\\[3pt]
72-713. NOTICE OF HEARINGS -{}- SERVICE. (1) The commission shall give at least ten (10) days' written notice \ldots{} either by personal service or by \sout{registered or certified mail} \uline{email or certified mail}.
}}
\end{center}
\vspace{-2pt}
\noindent{\footnotesize Source PDF text (Idaho HB 66, \S\,72-713): \sout{strikethrough} marks
deleted material, \uline{underline} marks inserted material.}
\vspace{4pt}

\begin{lstlisting}[style=snip,caption={Representative publication-XML fragment for the clause above; element names are stand-ins for the confidential schema vocabulary.},captionpos=b]
<section num="72-713">
  <heading>NOTICE OF HEARINGS -- SERVICE</heading>
  <para>The commission shall give at least ten (10)
    days' written notice ... either by personal
    service or by
    <deleted.material>registered or certified
      mail</deleted.material>
    <added.material>email or certified
      mail</added.material>.
    <cite.query resolved="ID-CODE-72-713">Section
      72-713, Idaho Code</cite.query>
  </para>
</section>
\end{lstlisting}

\paragraph{How the measures read this example.}
The four dimensions of \S\ref{sec:eval}---five measures, once citations are split---each score a
different part of the fragment above. \emph{Textual integrity} compares the running words of the
paragraph, ``The commission shall give at least ten (10) days' written notice \ldots{}'', against
the gold text character by character, and penalizes a dropped ``(10)'' or a changed ``ten''.
\emph{Layout topology} ignores those words and checks only the shape: a \texttt{section} holding a
\texttt{heading} and a \texttt{para}, in that order. \emph{Legislative mutations} checks the two
inline spans, requiring \texttt{deleted.material} to carry exactly ``registered or certified mail''
and \texttt{added.material} exactly ``email or certified mail''; swapping the two tags, or dropping
either span, counts as a miss even though every word is still present on the page.
\emph{Citation fidelity} splits in two here: localization asks whether ``Section 72-713, Idaho
Code'' was found and wrapped in a \texttt{cite.query} at all, and normalization asks whether the
resolved attribute carries the correct statutory identifier---the step no model completed unaided
(\S\ref{sec:citation}).

\section{Corpus Size Distribution}

Table~\ref{tab:corpus} gives the size of the 153-document gold set (\S\ref{sec:task}) broken down by jurisdiction, putting numbers to the ``skews small, long right tail'' shape noted there. Most documents are small, and two New Jersey budget bills sit at the far end.

\begin{table}[t]
  \centering
  \small
  \setlength{\tabcolsep}{2.4pt}
  \footnotesize
  \begin{tabular}{lccccccc}
    \toprule
    & & \multicolumn{3}{c}{Body nodes} & \multicolumn{3}{c}{Length (chars)} \\
    \cmidrule(lr){3-5}\cmidrule(lr){6-8}
    Cohort & $n$ & med & p90 & max & med & p90 & max \\
    \midrule
    All        & 153 & 130 & 652   & 40{,}944 & 5.8K & 42K & 1.16M \\
    Idaho      & 103 & 97  & 366   & 1{,}955  & 4.0K & 23K & 42K \\
    New Jersey & 50  & 187 & 1{,}744 & 40{,}944 & 9.4K & 74K & 1.16M \\
    \bottomrule
  \end{tabular}
  \caption{Corpus size distribution over the 153-document gold set (103 Idaho, 50 New Jersey). Documents vary in breadth, not depth. XML tree depth is $8$ for nearly all documents (max $9$), fixed by the shallow schema hierarchy, so we report body-node and character-length distributions. New Jersey is larger than Idaho at every percentile. ``Length'' is the character count of the bill body (K/M denote thousands/millions).}
  \label{tab:corpus}
\end{table}

\section{Production Pipeline and State Handling}
\label{app:arch}

\begin{figure*}[t]
  \centering
  \includegraphics[width=\textwidth]{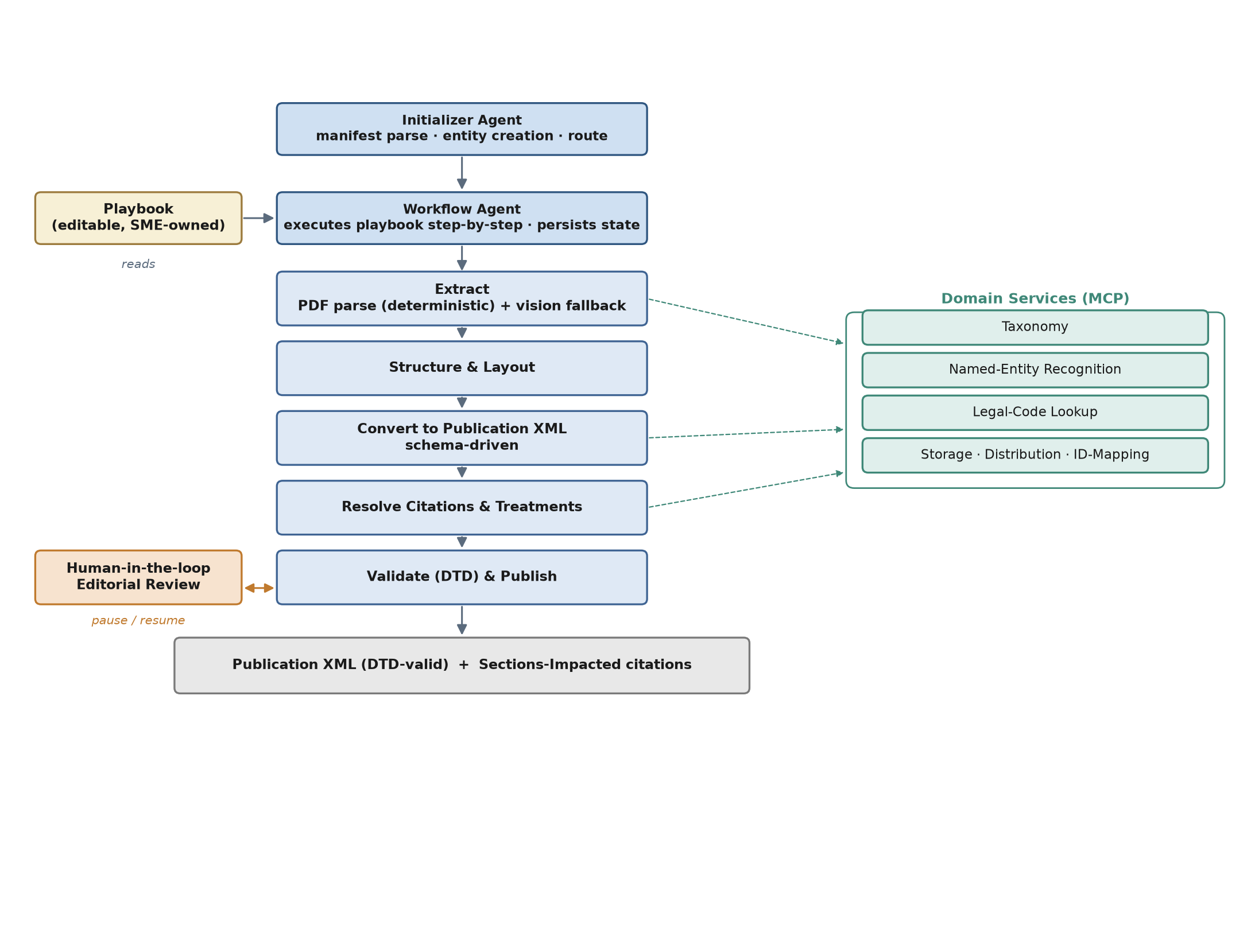}
  \caption{The T3 harness, modeled on the production pipeline (the most-scaffolded harness in Figure~\ref{fig:tiers}). The ablation runs it with a single uniform model across every agent, unlike the deployed system's mixed configuration (see the Limitations section). An initializer routes each input to the workflow agent, which reads an editable playbook and runs a six-stage task-agent spine (document extraction, structure and auxiliary-artifact construction, statute-impact extraction, XML conversion, citation resolution and footnoting, validation). The initializer and these six stage agents are the seven sub-agents counted in Table~\ref{tab:harness}. Stages call domain services for storage, taxonomy, ID-mapping, NER, codes lookup, and distribution, and editorial review gates publication after validation. T2 exposes this same tool set to a single agent, and T1 uses none of it.}
  \label{fig:arch}
\end{figure*}

\begin{figure*}[t]
  \centering
  \includegraphics[width=\textwidth]{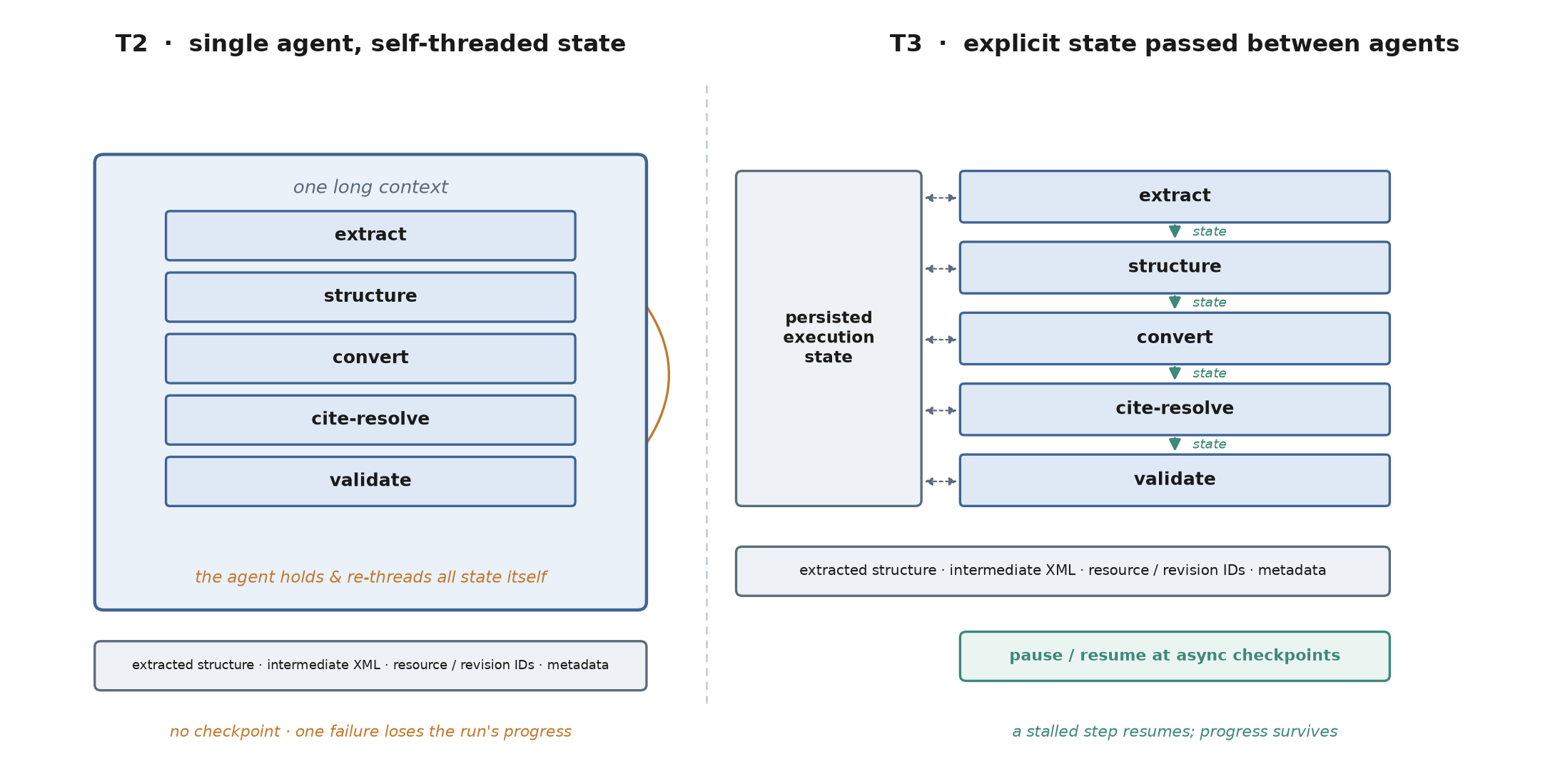}
  \caption{State handling in the two agentic harnesses. In T2 (left) a single agent holds and
  re-threads all intermediate state inside one long context with no checkpoint, so a failure
  loses the run's progress. In T3 (right) state is passed explicitly between agents through a
  persisted execution store, so a run can pause and resume at asynchronous checkpoints and a
  stalled step recovers.}
  \label{fig:state}
\end{figure*}

This section lays out the deployed T3 harness from \S\ref{sec:system}. It covers the domain services the agents call, the six-stage pipeline (Figure~\ref{fig:arch}), and the state-handling contrast between the two agentic harnesses (Figure~\ref{fig:state}) that \S\ref{sec:results} and the Limitations section point to.

\paragraph{Domain services.} The pipeline calls six MCP-backed services (\S\ref{sec:system}):
\begin{itemize}
  \item \emph{storage}, which handles state persistence, rendition reads and saves, and revision cloning.
  \item \emph{taxonomy}, which supplies jurisdiction, session, and measure-type metadata.
  \item \emph{ID-mapping}, which aligns identifiers across years and versions.
  \item \emph{NER}, for named-entity linking.
  \item a \emph{codes} service, whose \texttt{statute\_lookup\_batch} tool returns each citation's canonical, jurisdiction-specific identifier in one round-trip.
  \item \emph{distribution}, which hands the validated document to the publish step.
\end{itemize}

\section{Per-Dimension and Reliability Detail}
\label{app:detail}

\paragraph{Why \texttt{Overall} is not the row mean.}
Table~\ref{tab:main}'s \texttt{Overall} column is computed per document and then macro-averaged,
while each displayed column pools over a slightly different set of documents, so the two need not
agree. The gap is widest at zero-shot, where documents with no amendment spans---scored as true
negatives in the all-documents variant---are most common: naively averaging Haiku~4.5~T1's
displayed columns gives $0.594$ against the reported $0.677$. The denominator mismatch accounts
for every row's gap; none of it is a computation error.

\begin{figure*}[t]
  \centering
  \includegraphics[width=\textwidth]{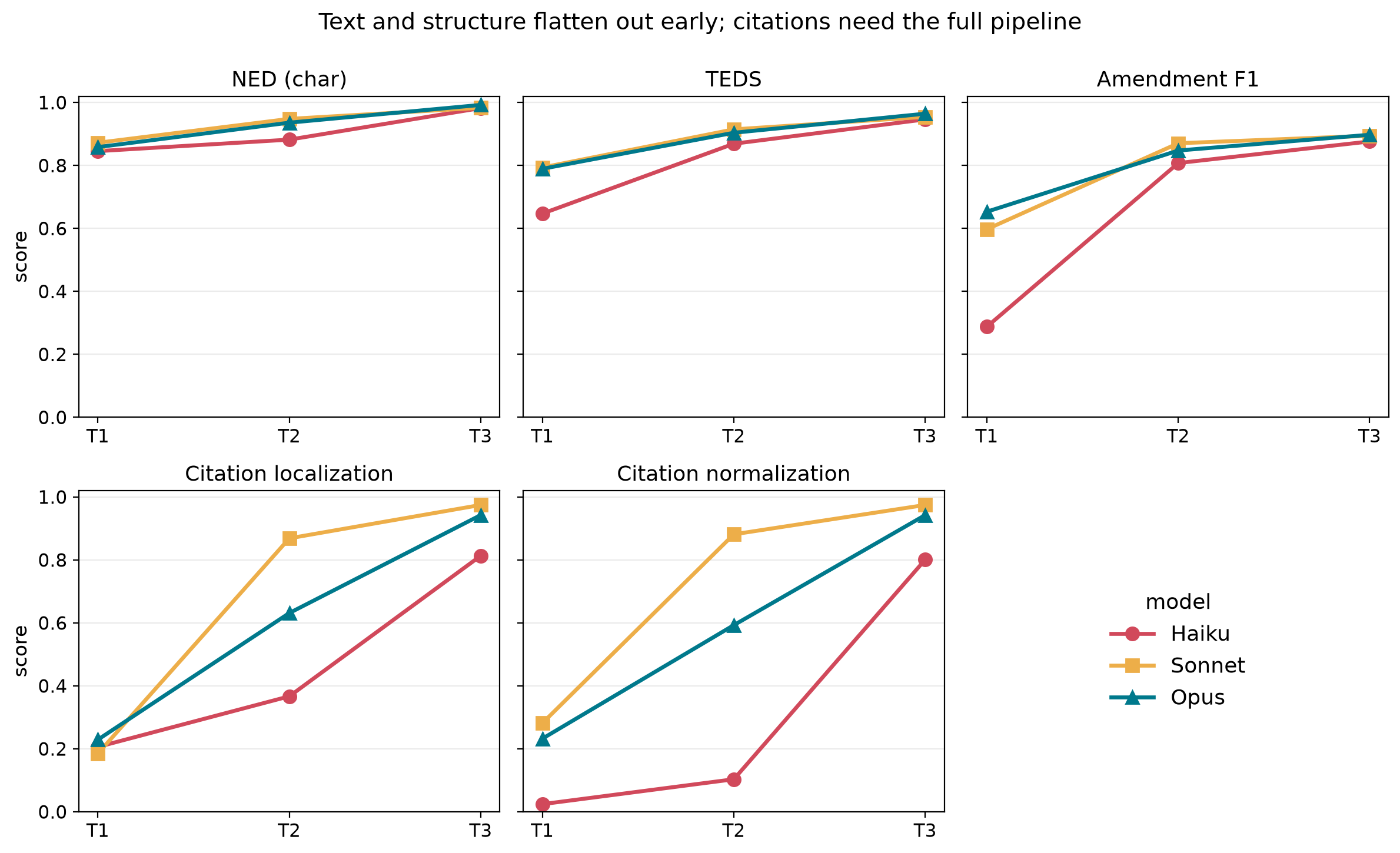}
  \caption{The five dimensions across the nine cells (exact values in Table~\ref{tab:main}).
  Text and structure (NED, TEDS) are already high at T1 and flatten quickly; amendments improve
  mostly on the T1$\rightarrow$T2 jump; citations are the one dimension that stays low until the
  full pipeline, and the one where Haiku visibly lags Sonnet and Opus.}
  \label{fig:dimensions}
\end{figure*}

\begin{figure*}[t]
  \centering
  \includegraphics[width=0.82\textwidth]{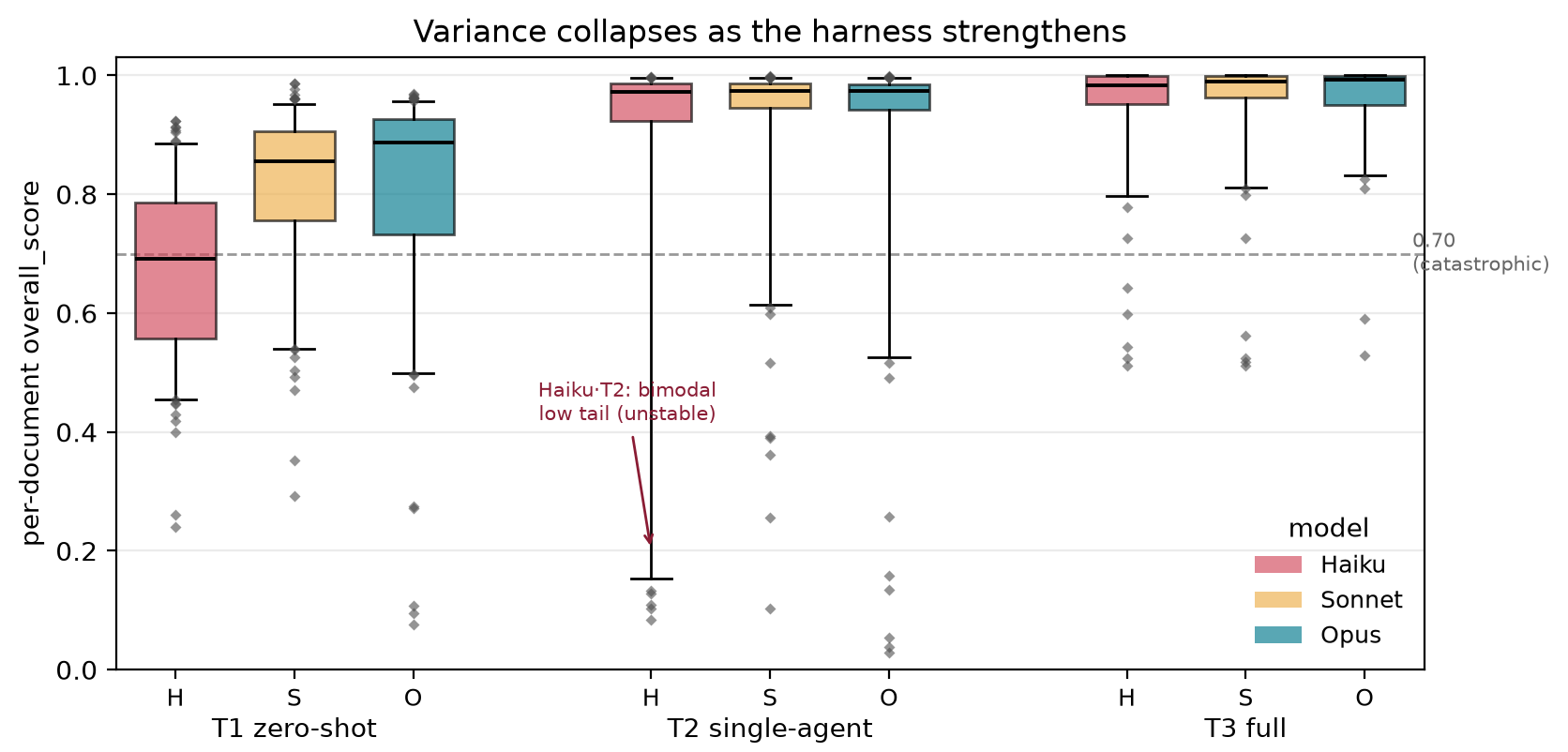}
  \caption{Per-document \texttt{Overall} score distribution across the nine configurations,
  the variability behind the pooled statistics in Table~\ref{tab:reliability}. Variance is
  non-monotonic: the single-agent harness (T2) is the widest, and only the full pipeline (T3)
  collapses both the spread and the low-scoring left tail.}
  \label{fig:reliability}
\end{figure*}

These two figures expand the aggregate results of \S\ref{sec:results}. Figure~\ref{fig:dimensions} breaks the composite into its five constituent dimensions, and Figure~\ref{fig:reliability} shows the per-document score distribution that sits behind the pooled reliability statistics in Table~\ref{tab:reliability}.

\section{Stress Bills and Citation-Residual Detail}

This section gives the per-document detail behind two results from \S\ref{sec:results}. The first is the collapse-and-recovery of the two largest New Jersey budget bills (Table~\ref{tab:stress}). The second is a qualitative taxonomy of the T3 citation-normalization residual (Table~\ref{tab:citetax}, \S\ref{sec:citation}).

\begin{table}[hb]
  \centering
  \small
  \setlength{\tabcolsep}{3.5pt}
  \begin{tabular}{llrrcc}
    \toprule
    Bill & Harness & Gold & Pred & NED & Amend.\ $F_1$ \\
    \midrule
    SB\_2025 & T1 & 40{,}944 & 1{,}087 & 0.10 & 0.00 \\
    SB\_2025 & T3 & 40{,}944 & 22{,}239 & \best{0.87} & \best{0.92} \\
    AB\_4700 & T1 & 8{,}229 & 334 & 0.12 & 0.00 \\
    AB\_4700 & T3 & 8{,}229 & 22{,}146 & \best{0.95} & \best{0.97} \\
    \bottomrule
  \end{tabular}
  \caption{Stress-bill collapse and recovery, T1 vs.\ T3, both scored on Sonnet, the only model with a completed run on these two documents (so the pattern is not verified to generalize to Haiku or Opus). TEDS is unscored under the node cap, but every scorable dimension moves from abandoned to solved. Predicted node counts exceed gold at T3 because the pipeline's serialization further subdivides interior structure, such as clause, citation, and amendment spans, into more atomic XML nodes than the gold annotation uses. NED and amendment $F_1$ score character- and span-level content rather than node count, so this inflation does not by itself depress those scores.}
  \label{tab:stress}
\end{table}

\begin{table}[hb]
  \centering
  \small
  \begin{tabular}{lcl}
    \toprule
    Failure mode & Docs & Example \\
    \midrule
    Dropped / partial subset & 23 & NJ SB~3587 \\
    Granularity / neighbor id & 3 & ID SB~1023 \\
    Lookup miss (left unresolved) & 2 & ID SB~1142 \\
    Range over-expansion & 1 & ID SB~1008 \\
    \midrule
    Localization text-capture artifact & 3 & NJ AB~5729 \\
    \bottomrule
  \end{tabular}
  \caption{Qualitative taxonomy of the T3 citation-normalization residual: the 32 of 167 outputs containing at least one \texttt{cite.query} node (pooled across the three T3 models) with any imperfect resolution, grouped by failure mode with a representative bill and assigned from a diff of predicted against gold \texttt{cite.query} resolutions (\S\ref{sec:citation}). The dropped-subset and lookup-miss modes are coverage-limited, meaning the codes service returns no identifier. The range over-expansion is a single Idaho outlier that on its own inflates that jurisdiction's raw residual. The final row is not a resolution error: the identifier resolves correctly, but the captured visible span differs from gold.}
  \label{tab:citetax}
\end{table}

\end{document}